\ifdefined\XeTeXversion\else\ifdefined\directlua\else\pdfoutput=1\fi\fi 
\documentclass[11pt]{article}
\usepackage[nologo]{thefinai}
\usepackage{adjustbox,latexsym,pifont,fvextra}
\DeclareCaptionFont{10pt}{\fontsize{10}{12}\selectfont}

\renewcommand{\cite}{\citep}
\graphicspath{{fig/}}

\newcommand{\cmark}{\textcolor{green}{\ding{51}}}
\newcommand{\xmark}{\textcolor{red}{\ding{55}}}
\newcommand{\pmark}{$\triangle$}

\title{Can LLMs Understand What They Predict? Evaluating Latent Causal Structure Inference in Financial Markets}
\date{}

\author{%
  Yixiang Zheng\textsuperscript{1}, 
  Yuechen Jiang\textsuperscript{1}, 
  Zhiwei Liu  \textsuperscript{1},
  Jimin Huang  \textsuperscript{1,2},
  Yuyan Wang\textsuperscript{1}, 
  Sophia Ananiadou\textsuperscript{1,3,\dag}%
}
\affiliation[1]{National Centre for Text Mining,
The University of Manchester}
\affiliation[2]{The Fin AI}
\affiliation[3]{ELLIS Manchester}
\authornote{\textsuperscript{\dag}Corresponding author}

\begin{document}

\begin{abstract}

Large language models (LLMs) are increasingly evaluated through their ability to predict future outcomes, implicitly assuming that predictive success reflects an understanding of the underlying system. However, accurate predictions can often be achieved without recovering the latent mechanisms that generate observable outcomes. To examine this assumption, we introduce \textsc{LUCID}, a benchmark that evaluates whether models can infer latent market structures from event descriptions and historical financial observations. Models generate event-induced causal graphs representing inferred market structure and are evaluated through a reconstruction-based framework. Our evaluation of 6 open-weight and proprietary LLMs shows that adding their generated graphs generally reduces reconstruction quality compared with using historical observations alone, although a few metrics improve. These findings suggest that inferring informative, event-specific market structures from observable evidence remains challenging for the evaluated LLMs.

\end{abstract}

\maketitle

\section{Introduction}

\begin{figure}[t]
\centering
\includegraphics[width=\columnwidth]{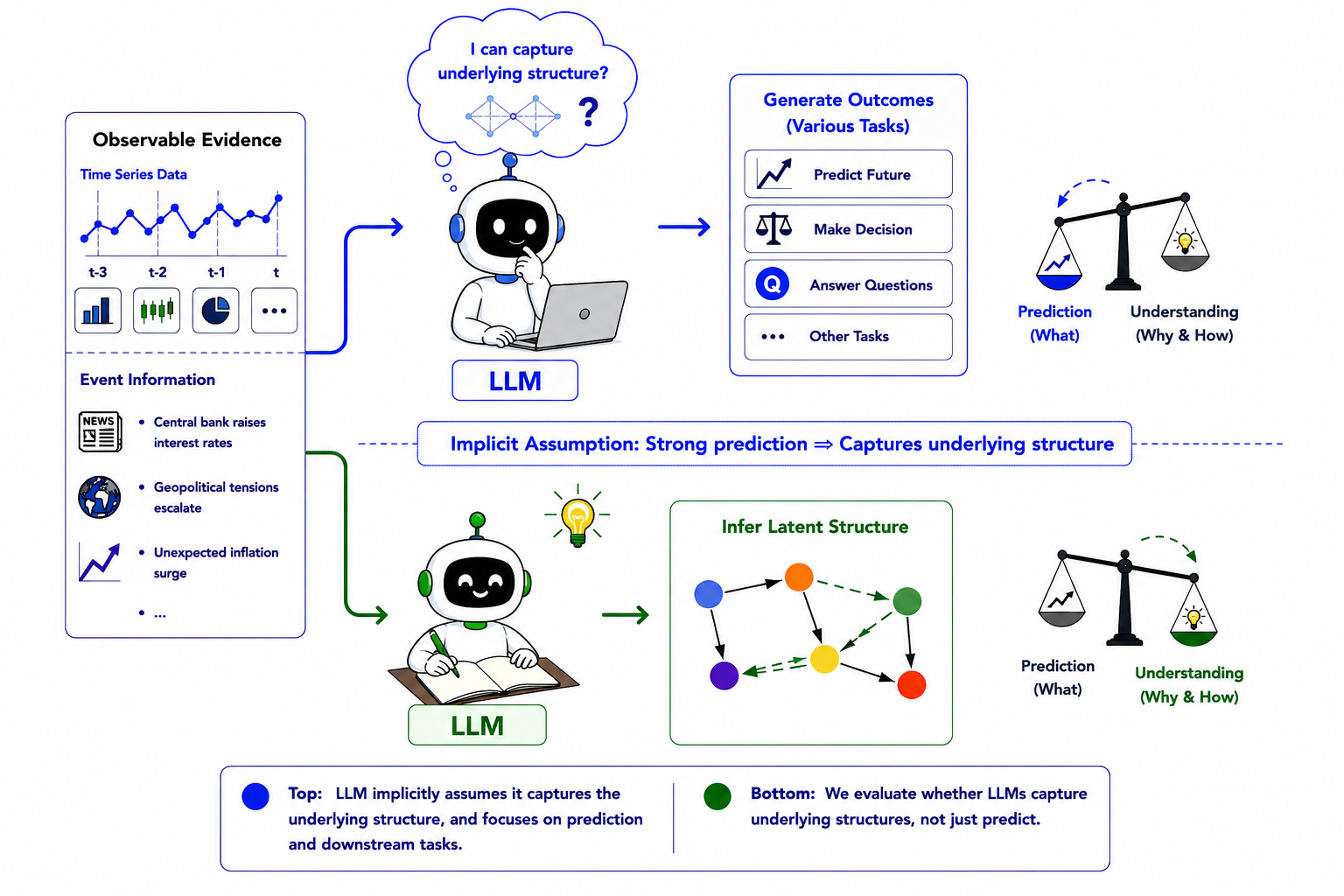}
\caption{
Prediction versus Understanding.
}
\label{fig:intro_comparison}
\end{figure}

The ability to accurately predict future observations is often treated as evidence of understanding, from scientific theories to modern large language models \citep{turing1950computing,binz2025centaur,binz2026posttraining}. Underlying this view is an implicit assumption: \textit{models that achieve strong predictive performance must also capture the underlying structure of the system being modeled} \citep{breiman2001statistical,pearl2009causality,lake2017building,rmus2026process}. However, this assumption has not been fully examined. A model may produce accurate predictions by exploiting statistical regularities, historical correlations, or recurring patterns without recovering the latent mechanisms that generate those observations \citep{breiman2001statistical,lapuschkin2019unmasking,geirhos2020shortcut,rmus2026process,binz2026posttraining}. This raises an important question: \textbf{can LLMs infer latent structures from observable evidence, rather than merely predicting observable outcomes?} (Figure~\ref{fig:intro_comparison}) \citep{pearl2009causality,granger1969investigating,scholkopf2021toward,mechanistic2026understanding}. This distinction is particularly salient in financial markets, where identical market outcomes may emerge from different underlying mechanisms and transmission pathways \citep{fama1965behavior,fama1970efficient,shiller1981stock,chen1986economic,cont2001empirical}. Yet existing evaluations of financial LLMs primarily assess forecasting accuracy, trading performance, or decision quality, offering limited insight into whether models can infer the latent structures underlying observable market dynamics \citep{campbell1997econometrics,cont2001empirical,finmem2023,fincon2024,xie2024finben, investorbench2025}.

\begin{table*}[t]
\centering
\fontsize{9pt}{10.8pt}\selectfont
\setlength{\tabcolsep}{3pt}

\begin{tabular*}{\textwidth}{@{\extracolsep{\fill}}llccc@{}}
\toprule
\textbf{Paradigm} &
\textbf{Evaluation Target} &
\shortstack{\textbf{Multimodal}\\\textbf{Evidence}} &
\shortstack{\textbf{Event-}\\\textbf{Conditioned}} &
\textbf{Explanation} \\
\midrule

LLMTIME \citep{gruver2023large}
& Forecasting & \xmark & \xmark & \xmark \\

CiK \citep{williams2024context}
& Forecasting & \cmark & \xmark & \xmark \\

\citep{merrill2024language}
& Forecasting/Question-Answering & \cmark & \xmark & \xmark \\

FinBen \citep{xie2024finben}
& Forecasting/Decision-Making & \cmark & \xmark & \xmark \\

\midrule

CLADDER \citep{jin2023cladder}
& Local Causal Relationship & \xmark & \xmark & \xmark \\

LLMGC \citep{liu2026llm}
& Graph Recovery & \pmark & \xmark & \xmark \\

\midrule

\textbf{LUCID(Ours)}
& Latent Structure Inferring & \cmark & \cmark & \cmark \\

\bottomrule
\end{tabular*}

\caption{
Comparison of representative paradigms for temporal forecasting, causal reasoning, and latent structure inference. We compare the evaluation target, the use of multimodal evidence, support for event-conditioned analysis, and the ability to provide explanatory reasoning. Symbols denote the level of support: \cmark indicate full support, \xmark indicates the absence of support, and \pmark indicates partial or limited support.
}
\label{tab:benchmark_comparison}
\end{table*}

Existing time-series evaluation paradigms assess LLM forecasting, time-series question answering, and financial decision making. For instance, \citet{gruver2023large} evaluates LLM forecasting ability directly from historical time-series observations, while \citet{williams2024context} incorporates textual information into the forecasting setting. Beyond forecasting, \citet{merrill2024language} assesses question answering over time series, and \citet{xie2024finben} evaluates financial decision making and other domain-specific capabilities. Causal reasoning benchmarks include \citet{jin2023cladder}, which evaluates local causal relationships using textual information, and \citet{liu2026llm}, which studies causal graph recovery using graph-internal knowledge such as variable semantics. Table~\ref{tab:benchmark_comparison} summarizes their evaluation targets and evidence types.

To bridge this gap, we propose \textsc{LUCID}, a benchmark for evaluating whether LLMs can infer latent structures from observable evidence. Specifically, models are given an event description and historical financial observations and are required to generate an event-induced causal graph representing the model’s inferred market structure. This formulation shifts the evaluation focus from outcome prediction to explicit structure inference. Because latent structures are not directly observable in real-world markets, we introduce a reconstruction-based evaluation framework that measures the incremental utility of generated graphs for market reconstruction.

We evaluate six open-weight and proprietary LLMs and find that, although graphs from some models improve individual metrics, no model jointly improves reconstruction fidelity and market-distribution consistency over its matched time-series-only baseline under our evaluation protocol. Further analyses reveal recurring relations and explanations across events in several models. Case studies further show that reused explanations can overlook event-specific context and that edge polarities can contradict their explanation. Together, these findings suggest that translating observable evidence into informative, event-specific market structures remains challenging for the evaluated LLMs.

\begin{itemize}

\item We propose \textsc{LUCID}, a benchmark for evaluating latent structure inference in financial markets, built from real-world events paired with historical financial observations. It evaluates whether LLMs can infer latent market structures from observable evidence.

\item We define an event-conditioned causal graph generation task that requires models to infer market relationships and transmission pathways and provide edge polarities and mechanism explanations, moving beyond conventional forecasting objectives.

\item We introduce a \textbf{reconstruction-based evaluation framework} that assesses the incremental utility of generated graphs for market reconstruction. Using matched time-series-only baselines and graph controls, it evaluates reconstruction fidelity and market-distribution consistency.

\end{itemize}

\begin{figure}[t]
\centering
\includegraphics[width=\textwidth]{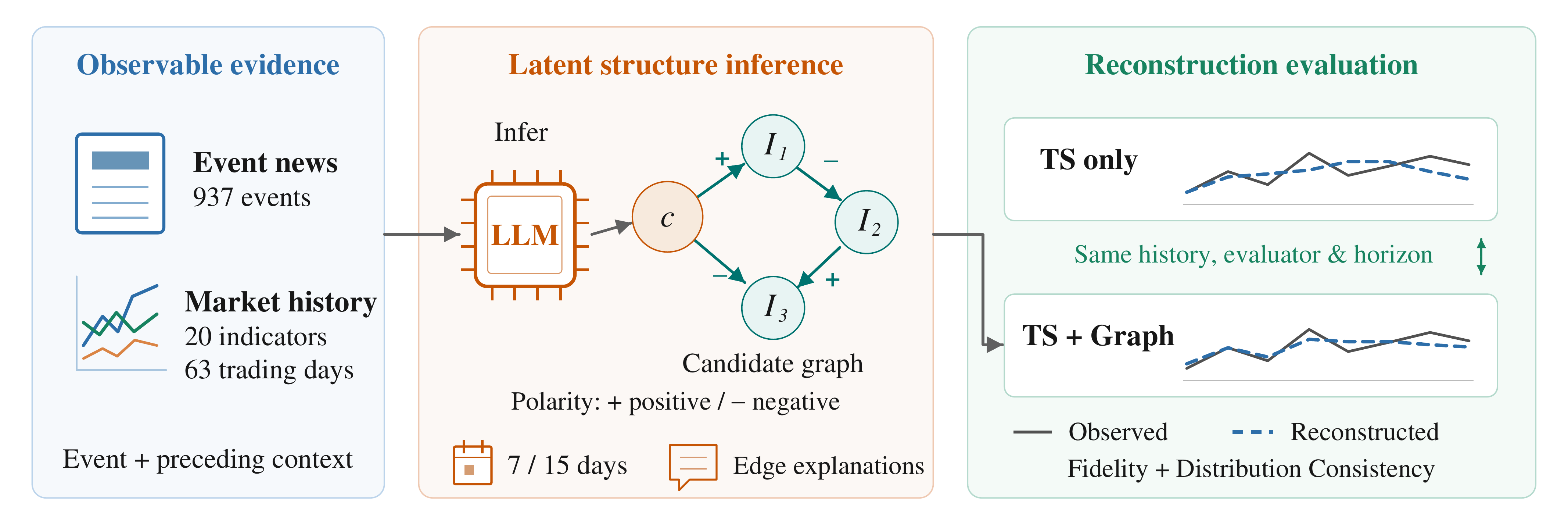}
\caption{
Overall workflow of \textsc{LUCID}.
}
\label{fig:overview}
\end{figure}

\section{Related Work}

Early studies have increasingly explored the use of LLMs for time-series forecasting \citep{gruver2023large, xue2023promptcast}. Building upon these advances, a growing body of work incorporates external textual events into multimodal forecasting systems, combining textual information with time-series observations to improve future outcome prediction \citep{jia2024gpt4mts, cao2023tempo, wang2024news}. Beyond forecasting, a growing body of work investigates the causal reasoning capabilities of LLMs, including causal relation identification \citep{wang2024causalbench}, counterfactual reasoning \citep{jin2023cladder}, and multi-hop causal chain reasoning \citep{romanou2023crab}. Existing causal reasoning benchmarks primarily evaluate local causal reasoning from textual information. In parallel, several studies use LLMs to assist causal discovery by providing semantic knowledge about variables and their relationships for historical structure recovery \citep{ban2023query,cohrs2024large}.

\section{\textsc{LUCID}}
Figure~\ref{fig:overview} summarizes the \textsc{LUCID}
workflow, from observable evidence and latent structure
inference to reconstruction-based evaluation.

\subsection{Task Formulation}

\textsc{LUCID} defines a causal graph generation task to evaluate whether LLMs can infer the latent structural regularities underlying financial markets from event descriptions and historical financial observations. The generated causal graph serves as an explicit representation
of the model's hypothesis about the latent market structure. Let $c \in \mathcal{C}$ denote an event summary, where $\mathcal{C}$ is the set of all event summaries. Let $t_0$ denote the timestamp of the event and $\mathcal{I}=\{I_1,\ldots,I_{20}\}$ denote the set of financial indicators. For each indicator $I_j$, we collect its historical observations during the 63 trading days preceding the event, denoted as $\mathbf{x}_j=\{x_j^{(t_0-62)},\ldots,x_j^{(t_0)}\}$. The complete financial context is therefore represented as $\mathbf{X}=\{\mathbf{x}_1,\ldots,\mathbf{x}_{20}\}$.

Given $(c,\mathbf{X})$, the model is required to generate an event-induced graph $G = (V, R, S, \Phi)$. Each node $v\in V$ represents the event or a financial indicator. A directed edge $(u,v)\in R$ represents a hypothesized direct effect from the event to an indicator or a subsequent transmission relation between indicators. Each edge is assigned a polarity
$s_{uv} \in \{\texttt{positive}, \texttt{negative}\}$, recorded in $S$. Positive and negative denote upward and downward direct effects, or propagation in the same and opposite directions as the source's event-induced change, respectively. Each edge is also accompanied by a mechanism explanation $\phi_{uv}\in\Phi$, describing how the event directly affects an indicator or how its effects propagate between indicators. Since the impacts of different events may persist for different periods, the model is additionally required to select an event duration $d \in \mathcal{D} = \{7, 15\}$, which defines the future time window over which the graph represents the event's market propagation. The task is formulated as:

\begin{equation}
(G^*, d^*) =
\underset{G \in \mathcal{G},\, d \in \mathcal{D}}{\arg\max}
P_{\mathrm{LLM}}(G,d \mid c,\mathbf{X}).
\end{equation}

where $\mathcal{G}$ denotes the space of valid event-induced graphs with edge polarities and mechanism explanations. and $G^*$ and $d^*$ are the graph and event duration predicted by the model, respectively.

\subsection{Data Curation}

Based on the task definition, we curate the \textsc{LUCID} using real-world financial events and historical market observations. We collect 937 real-world events from news articles retrieved through the Guardian Open Platform API.\footnote{The Guardian Open Platform API: \url{https://open-platform.theguardian.com/}. Original news content remains subject to the Guardian's copyright and licensing terms and is not redistributed as part of this benchmark. The benchmark annotations produced in this work are distributed under the Creative Commons Attribution License (CC BY 4.0).}
Related articles are grouped into event clusters, and each event is represented using the headline and trail text of the earliest published article in its cluster, providing a single event time and preventing later reports from entering the event description (the detailed collection, curation procedures, and temporal distribution of events are provided in Appendix~\ref{app:event_pipeline}). The event dataset spans the period from September 24, 2018 to December 18, 2025. 

For the observable market context, we collect 20 financial indicators, including market indices, exchange-traded funds (ETFs), from Yahoo Finance API.\footnote{Yahoo Finance: \url{https://finance.yahoo.com/}. All underlying data remain subject to the copyright, licensing, and terms of use of their original providers. Processed benchmark records released with this work are distributed under the Creative Commons Attribution License (CC BY 4.0).} The time-series dataset spans the period from January 2, 2015 to March 26, 2026. Each benchmark instance pairs an event with the preceding 63 trading-day histories of all indicators. The complete indicator set is reported in Appendix~\ref{app:data_statistics}.

\subsection{Evaluation}
\label{sec:evaluation-protocol}

We evaluate a diverse set of LLMs spanning open-source and closed-source families, as summarized in Table~\ref{tab:llm_models}. The open-weight models include Meta's Llama series with different parameter scales~\citep{llama31_techreport} and Alibaba's Qwen series~\citep{qwen3technicalreport}. The proprietary models include OpenAI's GPT-5~\citep{openai_gpt5}, and Google's Gemini-3.6-Flash~\citep{google2026gemini36flash}. All models follow a unified prompting protocol, with the prompt template and inference details provided in Appendix~\ref{app:graph-generation-prompts}.

\begin{table}[ht]
\centering
\begin{adjustbox}{max width=\linewidth}
\begin{tabular}{lc}
\toprule
\textbf{Model} & \textbf{Organization}\\
\midrule

\multicolumn{2}{c}{\textit{Proprietary Models}}\\
GPT-5 \citep{openai_gpt5} & OpenAI\\
Gemini-3.6-Flash \citep{google2026gemini36flash} & Google\\

\midrule
\multicolumn{2}{c}{\textit{Open Source Models}}\\
Qwen3-8B-Instruct \citep{qwen3technicalreport} & Alibaba\\
Qwen3-32B-Instruct \citep{qwen3technicalreport} & Alibaba\\
Llama-3.1-8B-Instruct \citep{llama31_techreport} & Meta\\
Llama-3.3-70B-Instruct \citep{llama31_techreport} & Meta\\

\bottomrule
\end{tabular}
\end{adjustbox}
\caption{Overview of selected proprietary and open-source LLMs used in our evaluation.}
\label{tab:llm_models}
\end{table}

Direct evaluation of generated causal graphs is challenging because the latent structural relationships governing real financial markets are not directly observable. We therefore adopt a forecasting-based evaluation protocol, in which we assess whether the inferred propagation structures provide useful information beyond the historical observations for forecasting future market dynamics over the model-selected duration. Specifically, we use Qwen3.8-27B\citep{qwen38} as a fixed zero-shot forecasting evaluator distinct from the graph-generating models, given its long-context support for multivariate
historical observations and graph inputs. For every event--model pair, the evaluator forecasts the same model-selected 7- or 15-day future window under two conditions: \textbf{TS only}, which receives the historical observations, and \textbf{TS + Graph}, which receives the same observations plus the generated graph. The graph input preserves topology and polarity over anonymous indicator indices, excluding event text, indicator metadata, and mechanism explanations to limit additional semantic cues. Within each paired comparison, the event, history, duration, evaluator, output format, inference settings, and scoring set are matched. Thus, the difference between conditions isolates the effect of exposing the evaluator to the generated graph under this protocol.

Results from both durations are aggregated for overall evaluation, and the same evaluation protocol is applied across all evaluated LLMs. Detailed prompt templates, inference settings, output parsing, and score aggregation procedures are provided in Appendix~\ref{app: evaluator-implementation}. We further assess the utility of the generated graphs
against node-shuffled and event-mismatched controls,
and examine the evaluator's ability to exploit
informative graph inputs using VAR-based statistical
references.

\paragraph{Reconstruction Fidelity}
Let $\mathcal{I}_G \subseteq \mathcal{I}$ denote the set of indicators appearing in a generated graph and let $T=d$ denote the model-selected forecasting window. For each indicator $I_j \in \mathcal{I}_G$ and time step $t$, we define the absolute reconstruction error as
\begin{equation}
e_j^{(t)}
=
\left|
\hat{y}_j^{(t)}-y_j^{(t)}
\right|,
\end{equation}
where $y_j^{(t)}$ and $\hat{y}_j^{(t)}$ are the observed and
predicted values, respectively. We normalize each error as
$d_j^{(t)}=e_j^{(t)}/s_j$, where $s_j>0$ is the fixed
indicator-specific reference standard deviation defined in
Appendix~\ref{app:evl_metric}. With $m=|\mathcal{I}_G|$,
we summarize the standardized errors using their mean
and standard deviation:

\begin{equation}
\mu_d =
\frac{1}{mT}
\sum_{I_j \in \mathcal{I}_G}
\sum_{t=1}^{T}
d_j^{(t)}.
\end{equation}

\begin{equation}
\sigma_d =
\sqrt{
\frac{1}{mT}
\sum_{I_j \in \mathcal{I}_G}
\sum_{t=1}^{T}
\left(d_j^{(t)}-\mu_d\right)^2
}.  
\end{equation}

A smaller $\mu_d$ indicates lower average standardized
reconstruction error, while a smaller $\sigma_d$ indicates
lower dispersion of these errors across indicators and
time steps.

\paragraph{Market Distribution Consistency}
Reconstruction fidelity measures local accuracy, while market-distribution consistency characterizes the statistical properties of reconstructed market dynamics. We evaluate distribution consistency based on log returns, which are widely used in quantitative finance due to their additive properties and relative scale invariance~\citep{fama1965behavior,cont2001empirical}. Specifically, we compute log returns from the observed and reconstructed trajectories and aggregate them across all 20 indicators and time steps. Following prior studies showing that financial returns exhibit heavy-tailed behavior that deviates from Gaussian assumptions \citep{fama1965behavior,cont2001empirical}, we model the market return distribution using a Laplace distribution. Let $(\mu,b)$ and $(\hat{\mu},\hat{b})$ denote the location and scale parameters of the observed and reconstructed return distributions, respectively.

We measure the differences between their distributional parameters as
\begin{equation}
D_\mu = |\mu-\hat{\mu}|,
\qquad
D_b = |b-\hat{b}|,
\end{equation}
and define the overall distributional distance as
\begin{equation}
D_{\mathrm{dist}}
=
\sqrt{D_\mu^2+D_b^2}.
\end{equation}
Lower values indicate greater consistency between the reconstructed and observed market-return distributions. Detailed definitions of the log returns, aggregated return sets, parameter estimation, and numerical processing are provided in Appendix~\ref{app:evl_metric}.

\section{Experiments}



\subsection{Main Results}
\label{sec:task1}


Table~\ref{tab:main_results} summarizes future
market dynamics reconstruction performance when
incorporating event-induced graphs generated by
the evaluated LLMs. To examine whether reconstruction
benefits from the inferred indicator relationships
and their correspondence to the event, we additionally evaluate node-shuffled and event-mismatched controls. The node-shuffled
control preserves the involved indicator set and
topology shape while permuting indicator identities
and randomizing edge polarities, whereas the
event-mismatched control uses a graph generated
by the same LLM for another event with the same
duration. Each cell reports the relative percentage
change from the corresponding historical-time-series-only
baseline (negative changes indicate improvements,
whereas positive changes indicate degradation),
together with a paired-bootstrap 95\% confidence
interval in brackets. The baseline and both controls
are evaluated using the same event-level duration
routing and, for reconstruction fidelity, the same
indicator set as the original generated graph.
Control generation and confidence interval
estimation are described in
Appendices~\ref{app:graph-controls}
and~\ref{app:paired-ci}, while
absolute scores and full paired results are
provided in Appendix~\ref{app: add-result}.

\begin{table}[t]
\centering
\fontsize{10pt}{11pt}\selectfont
\setlength{\tabcolsep}{3pt}
\renewcommand{\arraystretch}{1.0}

\begin{adjustbox}{max width=\textwidth}
\begin{tabular}{@{}llccccc@{}}
\toprule
& &
\multicolumn{2}{c}{\textbf{Reconstruction Fidelity}} &
\multicolumn{3}{c}{\textbf{Market Distribution Consistency}} \\
\cmidrule(lr){3-4}
\cmidrule(l){5-7}
\textbf{Category} &
\textbf{Model} &
$\mu_d$ (\%) &
$\sigma_d$ (\%) &
$D_{\mu}$ (\%) &
$D_b$ (\%) &
$D_{\mathrm{dist}}$ (\%) \\
\midrule

\multirow{4}{*}{Proprietary}
& \multirow{2}{*}{GPT-5}
& $+3.95$
& $+2.24$
& $-2.28$
& $\mathbf{-3.10}$
& $\mathbf{-2.40}$ \\*
&
& $[0.80,\,7.28]$
& $[-1.05,\,5.71]$
& $[-8.08,\,4.18]$
& $[-4.94,\,-1.26]$
& $[-4.00,\,-0.77]$ \\

& \multirow{2}{*}{Gemini-3.6-Flash}
& $+0.86$
& $\mathbf{-2.15}$
& $+3.31$
& $+6.51$
& $+6.12$ \\*
&
& $[-2.51,\,4.23]$
& $[-5.53,\,1.24]$
& $[-2.71,\,9.77]$
& $[4.62,\,8.39]$
& $[4.49,\,7.70]$ \\
\midrule

\multirow{8}{*}{Open-Weight}
& \multirow{2}{*}{Qwen3-8B}
& $+3.90$
& $+3.20$
& $+1.87$
& $+3.88$
& $+3.34$ \\*
&
& $[0.98,\,6.84]$
& $[-0.19,\,6.71]$
& $[-3.83,\,7.84]$
& $[2.06,\,5.64]$
& $[1.87,\,4.75]$ \\

& \multirow{2}{*}{Qwen3-32B}
& $+3.32$
& $+3.82$
& $+1.99$
& $+0.03$
& $+0.10$ \\*
&
& $[0.49,\,6.31]$
& $[0.51,\,7.20]$
& $[-3.88,\,7.99]$
& $[-1.80,\,1.78]$
& $[-1.45,\,1.59]$ \\

& \multirow{2}{*}{Llama-3.1-8B}
& $+4.51$
& $+6.46$
& $+4.45$
& $+3.23$
& $+3.65$ \\*
&
& $[2.11,\,7.07]$
& $[3.31,\,9.69]$
& $[-1.15,\,10.50]$
& $[1.37,\,4.99]$
& $[2.26,\,5.11]$ \\

& \multirow{2}{*}{Llama-3.3-70B}
& $\mathbf{-3.72}$
& $-2.01$
& $\mathbf{-2.39}$
& $+8.14$
& $+6.89$ \\*
&
& $[-6.44,\,-0.93]$
& $[-5.19,\,1.17]$
& $[-9.35,\,5.04]$
& $[6.09,\,10.19]$
& $[5.28,\,8.63]$ \\

\bottomrule
\end{tabular}
\end{adjustbox}

\captionsetup{font={10pt,rm}}
\caption{
Main results on event-induced latent structure
inference.
Each cell reports the relative percentage change
from the corresponding historical-time-series-only
baseline, with a paired-bootstrap 95\% confidence
interval in brackets.
Relative change is computed as
$\Delta = 100 \times
(\text{Graph-conditioned} - \text{Historical-only})/
\text{Historical-only}$.
Negative relative changes indicate improvements,
whereas positive changes indicate degradation.
Bold values denote the smallest relative change
in each column.
}
\label{tab:main_results}
\end{table}

\begin{table}[t]
\centering
\fontsize{10pt}{12pt}\selectfont
\setlength{\tabcolsep}{4pt}
\renewcommand{\arraystretch}{1.1}

\begin{tabular}{@{}llccc@{}}
\toprule
& &
\multicolumn{2}{c}{\textbf{Reconstruction Fidelity}} &
\shortstack{\textbf{Market Distribution}\\
            \textbf{Consistency}} \\
\cmidrule(lr){3-4}
\cmidrule(l){5-5}
\textbf{Model} &
\textbf{Condition} &
$\mu_d$ (\%) &
$\sigma_d$ (\%) &
$D_{\mathrm{dist}}$ (\%) \\
\midrule

\multirow{3}{*}{Llama-3.3-70B}
& Real
& $\mathbf{-3.72}\,[-6.44,\,-0.93]$
& $\mathbf{-2.01}\,[-5.19,\,+1.17]$
& $+6.89\,[+5.28,\,+8.63]$ \\

& Shuffle
& $+0.76\,[-2.08,\,+3.68]$
& $+0.55\,[-2.74,\,+3.77]$
& $\mathbf{+6.66}\,[+5.02,\,+8.40]$ \\

& Mismatch
& $-1.20\,[-4.12,\,+1.72]$
& $+0.53\,[-3.33,\,+4.41]$
& $+7.41\,[+5.62,\,+9.26]$ \\
\midrule

\multirow{3}{*}{Gemini-3.6-Flash}
& Real
& $+0.86\,[-2.51,\,+4.23]$
& $\mathbf{-2.15}\,[-5.53,\,+1.24]$
& $\mathbf{+6.12}\,[+4.49,\,+7.70]$ \\

& Shuffle
& $+3.26\,[+0.44,\,+6.26]$
& $+1.70\,[-1.56,\,+5.08]$
& $+6.40\,[+4.69,\,+8.18]$ \\

& Mismatch
& $\mathbf{-0.28}\,[-3.08,\,+2.65]$
& $+0.29\,[-2.98,\,+3.73]$
& $+6.33\,[+4.85,\,+7.86]$ \\
\midrule

\multirow{3}{*}{GPT-5}
& Real
& $+3.95\,[+0.80,\,+7.28]$
& $+2.24\,[-1.05,\,+5.71]$
& $\mathbf{-2.40}\,[-4.00,\,-0.77]$ \\

& Shuffle
& $+1.69\,[-0.97,\,+4.53]$
& $+2.44\,[-0.68,\,+5.85]$
& $-1.11\,[-2.75,\,+0.49]$ \\

& Mismatch
& $\mathbf{+1.68}\,[-1.05,\,+4.64]$
& $\mathbf{+0.82}\,[-2.33,\,+4.24]$
& $-1.73\,[-3.33,\,-0.14]$ \\

\bottomrule
\end{tabular}

\captionsetup{font={10pt,rm}}
\caption{
Relative changes (\%) from the matched TS-only baseline
for original graphs (Real) and their controls with a paired-bootstrap 95\% confidence
interval in brackets. Negative values indicate improvements.
Bold indicates the lowest mean across graph conditions
for each model and metric.
}
\label{tab:selected-graph-controls}
\end{table}

Across all six models, none of the generated graph
sets yields a joint improvement in reconstruction
fidelity and market distribution consistency. Although some models exhibit partial gains, such
as a $3.72\%$ reduction in $\mu_d$ for Llama-3.3-70B
and a $2.40\%$ reduction in $D_{\mathrm{dist}}$ for
GPT-5, most metrics deteriorate relative to the
matched time-series-only baselines. Model scale provides partial benefits within the Llama family. Compared with the 8B variant, which degrades both $\mu_d$ and $\sigma_d$, the 70B variant shows improvements in both metrics on average relative to the matched time-series-only
baseline. However, $D_{\mathrm{dist}}$ remains above baseline, indicating that these reconstruction gains do not
extend to overall market distribution consistency. The proprietary models also exhibit different patterns
of partial utility. Gemini-3.6-Flash reduces the mean $\sigma_d$ but increases $D_{\mathrm{dist}}$. In contrast, GPT-5 improves $D_{\mathrm{dist}}$ while degrading $\mu_d$. These findings suggest that the evaluated
LLMs still face difficulty inferring event-induced
market structures that consistently provide
complementary information for reconstructing
individual indicator trajectories and aggregate
market dynamics.

Table~\ref{tab:selected-graph-controls} compares
original graphs with node-shuffled and event-mismatched
controls for models exhibiting partial gains.
The results provide localized evidence that preserving
graph relationships and event alignment can benefit
reconstruction. This is most evident for Llama-3.3-70B's $\mu_d$, where the original graphs outperform both controls with paired-bootstrap 95\% confidence intervals that exclude zero.

\begin{figure}[!t]
    \centering

    \begin{subfigure}[t]{0.48\linewidth}
        \centering
        \includegraphics[width=\linewidth]
        {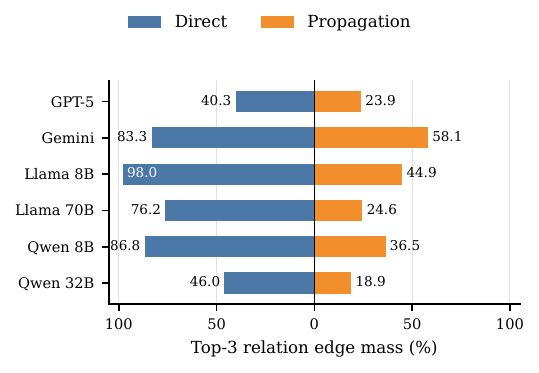}
        \caption{Structural concentration.}
        \label{fig:structural-concentration}
    \end{subfigure}
    \hfill
    \begin{subfigure}[t]{0.48\linewidth}
        \centering
        \includegraphics[width=\linewidth]
        {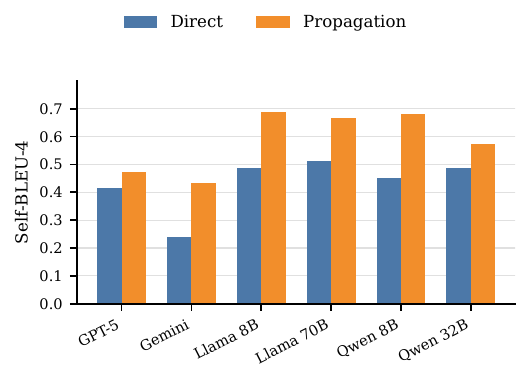}
        \caption{Explanation similarity.}
        \label{fig:explanation-similarity}
    \end{subfigure}

    \caption{
    Complementary analyses of generated graph structures and edge
    explanations across events. Direct and propagation edges are analyzed
    separately.
    }
    \label{fig:graph-diagnostics}
\end{figure}

\subsection{Evaluator Calibration with VAR References}

\begin{table}[t]
\centering
\fontsize{9pt}{10.8pt}\selectfont
\setlength{\tabcolsep}{3pt}
\renewcommand{\arraystretch}{1.2}

\newcommand{\varcalcell}[3]{%
\shortstack{$#1$\\{\scriptsize $[#2,#3]$}}%
}

\begin{tabular}{@{}>{\raggedright\arraybackslash}m{\dimexpr0.28\linewidth-4.5pt\relax}*{3}{>{\centering\arraybackslash}m{\dimexpr0.24\linewidth-4.5pt\relax}}@{}}
\toprule
&
\multicolumn{2}{c}{
\textbf{\shortstack{Reconstruction\\Fidelity}}}
&
\textbf{\shortstack{Market\\Distribution\\Consistency}}
\\
\cmidrule(lr){2-3}
\cmidrule(lr){4-4}

\textbf{Graph input}
& $\Delta\mu_d$ (\%)
& $\Delta\sigma_d$ (\%)
& $\Delta D_{\mathrm{dist}}$ (\%)
\\
\midrule

VAR
& \varcalcell{-13.56}{-20.61}{-5.85}
& \varcalcell{-9.42}{-16.42}{-1.51}
& \varcalcell{-4.72}{-6.59}{-2.74}
\\

Node-shuffled
& \varcalcell{+12.09}{+2.63}{+22.92}
& \varcalcell{+19.15}{+8.62}{+30.86}
& \varcalcell{+0.74}{-1.72}{+3.42}
\\

Event-mismatched
& \varcalcell{-0.86}{-6.85}{+5.54}
& \varcalcell{+0.01}{-6.25}{+6.47}
& \varcalcell{+1.46}{-1.55}{+4.31}
\\

\bottomrule
\end{tabular}

\caption{
Evaluator calibration with VAR reference graphs.
Each cell reports the relative percentage change
from the matched TS-only baseline, with a paired-bootstrap 95\% confidence interval in brackets. Negative values indicate improvements. Absolute scores and complete results for all five
metrics are provided in
Appendix~\ref{app: add-result}.
}
\label{tab:var-calibration}
\end{table}

We further examine whether the fixed forecasting evaluator can benefit from informative graph inputs. Specifically, we use VAR-based hindsight reference
graphs constructed from observed post-event trajectories. These graphs encode statistical relationships in realized market responses and
serve as calibration references.

We construct the reference graphs in three steps:
(1) fitting a VAR model ~\citep{pamfil2020dynotears} for each event using its
pre-event returns, with the VAR predictions serving
as a baseline for measuring post-event deviations ~\citep{mackinlay1997event};
(2) estimating cross-indicator lagged dependencies
using a regularized linear model~\citep{barigozzi2019nets}
fitted to post-event deviations from a separate set
of events excluded from evaluation; and (3) for each evaluation event, selecting direct-edge
targets from deviations on the first post-event day
and propagation edges using the lagged dependencies
estimated in step (2) and assigning polarities from the signs of first-day deviations and estimated coefficients
for direct and propagation edges, respectively.
We evaluate reference-graph utility relative to
the matched TS-only baseline and compare the
reference graphs with node-shuffled and
event-mismatched controls, following the main
evaluation protocol. Detailed reference-graph construction and control generation are provided in Appendix~\ref{app:var-calibration}.

As shown in Table~\ref{tab:var-calibration}, the VAR reference graphs reduce all three metrics relative
to the matched TS-only baseline, with the largest
reductions of 13.56\% and 9.42\% in $\mu_d$ and
$\sigma_d$, respectively. When these reference
graphs are replaced with node-shuffled or
event-mismatched controls while keeping the
historical observations fixed, performance deteriorates across all three metrics
relative to the VAR-reference condition.
The gains over TS only and their reduction under
the controls provide evidence that the evaluator
can use informative graph inputs to improve its
forecasts. This calibration therefore supports using Qwen3.8-27B as the evaluator to assess  whether LLM-generated propagation
structures provide useful forecasting information
beyond historical observations.

\subsection{Graph Specificity and Explanation Similarity}
\label{sec:graph-specificity}

To further characterize the generated graphs beyond their forecasting utility, we conduct two complementary analyses of direct event-to-indicator edges and indicator-to-indicator propagation edges across events: (1) structural concentration, measured as the combined proportion of the three most frequent relations among all edges of the corresponding type; and (2) explanation similarity, measured by Self-BLEU-4 \citep{zhu2018texygen} for explanations of the same relation across events, averaged over relations. Higher Self-BLEU indicates greater textual similarity and lower cross-event lexical diversity. Detailed definitions and complete results are provided in Appendix~\ref{app:graph-specificity}. In the main analysis, relations are grouped by their directed source--target pairs without distinguishing polarity.


Figure~\ref{fig:structural-concentration} shows substantial concentration in relation selection for several models. For Llama-3.1-8B, the three most frequent relations account for 97.97\% of direct edges and 44.93\% of propagation edges, showing that a small set of market channels dominates its direct-effect predictions across events. Qwen3-8B also exhibits high direct-relation concentration (86.81\%). By contrast, the larger variants in both the  Llama and Qwen families exhibit lower concentration for both edge types, indicating more dispersed relation selection. As shown in Figure~\ref{fig:explanation-similarity},
relation-conditioned Self-BLEU is generally higher
for open-weight than proprietary models and consistently higher for propagation than direct-effect explanations across all evaluated LLMs. These patterns indicate greater cross-event similarity in wording for the same relation among open-weight models and in propagation explanations. To determine whether this similarity includes reuse of identical wording, we additionally measure exact explanation reuse after normalizing case, punctuation, and whitespace. For Llama-3.1-8B, reuse reaches 58.13\% for direct effects
and 64.64\% for propagation relations, showing that its
textual similarity includes substantial verbatim reuse.
In contrast, Gemini exhibits only 0.05\% and 5.04\% reuse, respectively, despite its concentrated relation selection. This contrast shows that structural concentration and explanation reuse capture distinct output properties. Representative cases and detailed computation of exact explanation reuse are provided in Appendices~\ref{app: CS} and ~\ref{app:graph-specificity}. Taken together, these results reveal distinct patterns of relation concentration and explanation reuse across models. Some models repeatedly select a narrow set of market channels and reuse the same mechanism statements, whereas concentrated relation selection can also coexist with varied wording.

\begin{table}[t]
\centering
\fontsize{10pt}{12pt}\selectfont
\setlength{\tabcolsep}{4pt}
\renewcommand{\arraystretch}{1.0}

\begin{tabular*}{\columnwidth}{
@{\extracolsep{\fill}}lrr@{}
}
\toprule
\textbf{Model} &
\textbf{Direct} &
\textbf{Propagation} \\
\midrule
GPT-5
& 12.50 & 13.36 \\
Gemini-3.6-Flash
& 0.05 & 5.04 \\
Llama-3.1-8B
& \textbf{58.13} & \textbf{64.64} \\
Llama-3.3-70B
& 42.71 & 34.42 \\
Qwen3-8B
& 17.76 & 28.05 \\
Qwen3-32B
& 7.07 & 7.32 \\
\bottomrule
\end{tabular*}

\captionsetup{font={10pt,rm}}
\caption{
Exact explanation reuse (\%). Higher values indicate more frequent
reuse of identical explanation text for the same relation across events.
}
\label{tab:exact-explanation-reuse}
\end{table}

\subsection{Error Analysis}


To better understand the limitations of current LLMs in event-conditioned latent structure inference, we examine two types of local explanation deficiencies through selected case studies: insufficient adaptation to event context and inconsistency between edge polarity and its textual justification.

\paragraph{Insufficient contextual adaptation}
Llama-3.1-8B reuses ``Investors seek safe-haven assets, increasing gold demand'' for EVENT$\rightarrow$GLD across different events. For two of these events, however, expert analyses highlight distinct mechanisms that can weaken or offset this safe-haven effect (Appendix~\ref{app:contextual-adaptation}). The reused explanation omits these differences, offering a generic rationale without sufficiently justifying its application to each event.

\paragraph{Polarity--explanation inconsistency}
For the event ``Jeremy Hunt refuses to rule out sanctions against China'', Llama-3.1-8B writes ``US dollar strengthens as investors seek safe-haven assets,'' but assigns negative polarity to EVENT$\rightarrow$UUP, encoding dollar weakening. This case shows that a plausible mechanism description does not necessarily provide a consistent justification for the graph's predicted directional effect. More verified cases appear in Appendix~\ref{app:polarity-inconsistency}.

These cases highlight limitations in producing event-specific and internally coherent graph outputs: reused explanations may overlook relevant event context, while predicted polarities may contradict their stated mechanisms. Such deficiencies  raise further concerns about the reliability of the generated graphs as forecasting inputs.

\section{Conclusion and Future Work}

In this paper, we introduced \textsc{LUCID}, the benchmark for evaluating whether LLMs can infer latent structures in financial systems from event descriptions and historical financial observations. As latent structures in real financial systems are not directly observable, we further propose a reconstruction-based evaluation framework that assesses the incremental reconstruction utility of graph topology and polarity, complemented by analyses of relation selection and mechanism explanations.

\textsc{LUCID} makes inferred market mechanisms inspectable through explicit relations, edge polarities, and textual justifications. Combining reconstruction utility with analyses of relation selection and explanation consistency links mechanism interpretability to observable outcomes. The framework provides complementary diagnostics for event specificity, directional consistency, and explanatory coherence. Together, these analyses connect inspectable structural hypotheses with their measurable contribution to market reconstruction, supporting systematic comparison of mechanism-oriented reasoning across LLMs.

\textsc{LUCID} provides a testbed for developing and evaluating latent structure inference and system-level reasoning in LLMs. Future work should examine alternative representations, including probabilistic graphical models and agent-based interaction models, to capture uncertainty and interactions among market variables. Another direction is to develop inference methods that jointly enforce event-specific topology, edge polarity, and consistency with mechanism explanations, and assess their utility through matched reconstruction experiments and graph controls. 

\clearpage

\section*{Ethical Considerations}

\textsc{LUCID} is constructed from publicly available financial news and market data collected in accordance with the respective terms of use of the original data providers. To comply with licensing restrictions, the benchmark does not redistribute original news articles. Because edge explanations are generated by LLMs, they may
inherit inaccuracies, framing biases, or omissions from both the source
reporting and the generating models. We therefore identify these texts as
model-generated artifacts and do not treat them as authoritative descriptions
of financial events or mechanisms.

The benchmark is intended solely for evaluating latent structure inference in large language models rather than providing investment recommendations or supporting financial decision-making. Although it contains real-world financial events and market observations, the generated causal graphs should not be interpreted as validated causal mechanisms or actionable investment advice. Financial markets are influenced by numerous latent factors that are not directly observable, and multiple plausible mechanisms may explain the same observed market behavior. In addition, the benchmark should not be viewed as establishing the correctness of any inferred causal graph. Different language models may generate different yet plausible structural explanations due to differences in their pretraining data, reasoning processes, and inductive biases. Consequently, the generated graphs should be interpreted as hypotheses about possible market structures rather than verified representations of the underlying financial system. We hope the benchmark encourages systematic evaluation of structural reasoning while discouraging over-interpretation of generated causal explanations.

\section*{Limitations}

The proposed benchmark focuses on latent structure inference in financial markets and therefore evaluates only one class of latent reasoning problems. While financial markets provide a naturally occurring environment in which observable outcomes emerge from partially unobservable interactions, it remains an open question whether the findings of this benchmark generalize to other domains such as scientific discovery, healthcare, biological systems, or social reasoning. 

A second limitation concerns the intrinsic non-identifiability of latent structures. In complex real-world systems, multiple latent mechanisms may produce similar observable outcomes, making the inverse problem fundamentally underdetermined. Consequently, the benchmark does not aim to recover a unique ground-truth causal graph, but rather evaluates whether the inferred structures contain useful information for explaining and reconstructing observed market dynamics.

Furthermore, the proposed evaluation protocol assesses latent structures indirectly through graph-conditioned market reconstruction. Reconstruction quality should therefore be interpreted as a measure of \emph{structural utility} rather than direct verification of graph correctness. A graph that provides stronger structural constraints is expected to better support downstream reconstruction, but high reconstruction performance alone does not guarantee that the inferred graph faithfully represents the true underlying mechanism.

Finally, the quantitative evaluation depends on the capability of the fixed
forecasting evaluator and the prompting protocol used to incorporate the
generated graphs. A failure to improve forecasting may reflect either limited
structural utility in the graph or limitations in the evaluator's ability to
use graph information. The matched comparisons and VAR-reference calibration are intended to address this concern, but they do not eliminate evaluator dependence.

\section*{Acknowledgment}

The authors acknowledge The NaCTeM team and The Fin AI community for their research support, feedback, and collaborative
environment that contributed to this work. 
\bibliographystyle{acl_natbib}
\bibliography{references}

\appendix



\section{Dataset Statistics}
\label{app:data_statistics}

The financial indicator set consists of heterogeneous variables spanning multiple domains of the financial system. These indicators are grouped into categories for interpretability and coverage, but no structural or hierarchical constraints are imposed on their relationships. The 20 indicators and their corresponding metadata descriptions are listed below.

\paragraph{Risk and Asset Indicators}
\begin{itemize}
    \item \textbf{VIX}: CBOE Volatility Index, a proxy for market risk and sentiment.
    \item \textbf{GLD}: Gold ETF, representing safe-haven demand.
    \item \textbf{USO}: United States Oil Fund, capturing oil price dynamics.
    \item \textbf{UUP}: U.S. Dollar Index ETF, reflecting currency strength.
\end{itemize}

\paragraph{Sector Equity Indicators (ETFs)}
\begin{itemize}
    \item \textbf{XLB}: Materials sector (raw materials and commodities-related firms)
    \item \textbf{XLE}: Energy sector (oil, gas, and energy production)
    \item \textbf{XLF}: Financial sector (banks, insurance, and financial services)
    \item \textbf{XLI}: Industrials sector (manufacturing, transportation, and infrastructure)
    \item \textbf{XLK}: Technology sector (information technology and software)
    \item \textbf{XLP}: Consumer staples sector (essential consumer goods)
    \item \textbf{XLU}: Utilities sector (electric, gas, and water utilities)
    \item \textbf{XLV}: Health care sector (pharmaceuticals and medical services)
    \item \textbf{XLY}: Consumer discretionary sector (retail, leisure, and non-essential consumption)
    \item \textbf{XLRE}: Real estate sector (REITs and real estate services)
    \item \textbf{XLC}: Communication services sector (media, telecom, and internet platforms)
\end{itemize}

\paragraph{Broad Market Indices}
\begin{itemize}
    \item \textbf{SPY}: S\&P 500 ETF
    \item \textbf{QQQ}: Nasdaq-100 ETF
    \item \textbf{DIA}: Dow Jones Industrial Average ETF
\end{itemize}

\paragraph{Fixed Income and Interest Rate Indicators}
\begin{itemize}
    \item \textbf{TLT}: Long-term U.S. Treasury bond ETF (20+ years)
    \item \textbf{SHY}: Short-term U.S. Treasury bond ETF (1--3 years)
\end{itemize}


The indicator set is designed to provide broad coverage of financial scenarios and economic mechanisms, enabling the modeling of event-driven propagation across heterogeneous variables. Dependencies are not restricted by category, and causal relations may arise between any pair of indicators. This design avoids imposing inductive biases on the causal structure, allowing the benchmark to evaluate models' ability to discover cross-domain dependencies.



\section{Event Collection and Curation}
\label{app:event_pipeline}


We curate a collection of real-world external shock events. Among the collected events, most can be categorized into three groups: (i) \textit{geopolitical and political events}, (ii) \textit{bank crisis events}, and (iii) \textit{macroeconomic policy events}.

\paragraph{Retrieval and filtering}
We retrieve Guardian articles published between 2018 and 2025 using keywords related to major real-world events. After URL- and text-based deduplication, we remove missing records and non-event content, including opinion pieces, analyses, explainers, live blogs, newsletters, and reader callouts. This process retains 16,601 articles from 59,449 candidates.

\paragraph{Clustering and ranking}
We represent each article using its headline and trail text and encode the text with \texttt{all-MiniLM-L6-v2}. Articles are processed in monthly batches and grouped using hierarchical clustering with a cosine-similarity threshold of 0.70. Clusters spanning more than seven days are divided chronologically. After automatic relevance filtering, we prioritize clusters containing more articles and sample 1375 events
across months to maintain temporal coverage. Each event is represented
by the headline and trail text of the earliest published article in its
cluster.

\paragraph{Output validation and evaluation-set construction}
Starting from 1375 candidate events, we excluded 412 unique
events with invalid generated graphs, including directed
cycles, self-loops, duplicate edges, and violations of the
prescribed propagation-layer constraints. Forecasting
evaluation excluded a further 26 unique events whose outputs
reached the generation-length limit without yielding a
parseable forecast. Exclusions were deduplicated by event ID
across models and evaluation conditions, leaving 937
jointly valid events.

\paragraph{Event distribution}
The final dataset contains 937 events covering 88 consecutive months
from September 24, 2018 to December 18, 2025. Table~\ref{tab:event_year}
reports their yearly distribution.

\begin{table}[t]
\centering
\fontsize{11pt}{13.2pt}\selectfont
\begin{tabular*}{\columnwidth}{@{\extracolsep{\fill}}lrr@{}}
\toprule
Year & Number of events & Percentage \\
\midrule
2018 & 38  & 4.1\%  \\
2019 & 144 & 15.4\% \\
2020 & 132 & 14.1\% \\
2021 & 139 & 14.8\% \\
2022 & 107 & 11.4\% \\
2023 & 146 & 15.6\% \\
2024 & 117 & 12.5\% \\
2025 & 114 & 12.2\% \\
\midrule
Total & 937 & 100.0\% \\
\bottomrule
\end{tabular*}
\caption{Yearly distribution of real-world events, from
September 24, 2018 to December 18, 2025.}
\label{tab:event_year}
\end{table}

\section{Case Studies}
\label{app: CS}

\subsection{Insufficient Contextual Adaptation}
\label{app:contextual-adaptation}


Table~\ref{tab:reuse-context-cases} compares a reused explanation with financial expert analyses of three events.\footnote{The expert-analysis summaries are our paraphrases of articles published by
\href{https://www.marketpulse.com/news/oil-and-gold-dip-as-ukraine-crisis-eases/}{MarketPulse/OANDA},
\href{https://www.financialexpress.com/market/gold-set-for-biggest-weekly-drop-in-nearly-7-years-in-virus-led-sell-off/1896653/lite/}{Reuters (via The Financial Express)},
the \href{https://www.gold.org/goldhub/research/gold-prices-swing-as-markets-sell-off}{World Gold Council},
and \href{https://www.morganstanley.com/campaigns/wealth-management/move-report/move-report2023mar21/}{Morgan Stanley}.
Copyright in the original articles remains with the respective rights holders.}
The final sentence in each row states our assessment of the omitted context.

\nocite{erlam2022withdrawal,reuters2020liquidation,wgc2020liquidation,morganstanley2023yellen}

\begin{table}[t]
\centering
\small
\setlength{\tabcolsep}{4pt}
\renewcommand{\arraystretch}{1.15}
\begin{tabular}{@{}>{\raggedright\arraybackslash}p{0.30\textwidth}
                  >{\raggedright\arraybackslash}p{0.23\textwidth}
                  >{\raggedright\arraybackslash}p{\dimexpr0.47\textwidth-16pt\relax}@{}}
\toprule
\multicolumn{3}{@{}p{\textwidth}@{}}{\textbf{Shared model output:} Llama-3.1-8B; EVENT$\rightarrow$GLD; positive polarity; seven-day duration.} \\
\multicolumn{3}{@{}p{\textwidth}@{}}{\textbf{Reused explanation:} ``Investors seek safe-haven assets, increasing gold demand'' (87 distinct events after text normalization).} \\
\midrule
\textbf{Event and date} & \textbf{Expert source} & \textbf{Expert analysis and omitted context} \\
\midrule
\textbf{A. Partial troop withdrawal}\newline
15 February 2022. Russia announces a partial withdrawal from the Ukrainian border; western officials report no immediate evidence of drawdown.
& \href{https://www.marketpulse.com/news/oil-and-gold-dip-as-ukraine-crisis-eases/}{Erlam (2022)}\newline
OANDA / MarketPulse\newline
15 February 2022
& Attributes cooling gold demand to the withdrawal announcement's de-escalation signal. The reused explanation addresses neither this easing signal nor uncertainty about the withdrawal's credibility.
\\
\midrule
\textbf{B. Market sell-off and liquidity intervention}\newline
12 March 2020. The Fed offers \$1.5 trillion in liquidity as the Dow and FTSE suffer their worst day since 1987.
& \href{https://www.financialexpress.com/market/gold-set-for-biggest-weekly-drop-in-nearly-7-years-in-virus-led-sell-off/1896653/lite/}{Reuters (2020)}, quoting ANZ analyst Daniel Hynes\newline
13 March 2020\newline
\href{https://www.gold.org/goldhub/research/gold-prices-swing-as-markets-sell-off}{World Gold Council (2020)}\newline
19 March 2020
& Discusses broad liquidation and gold sales to raise cash and meet margin calls during the sell-off. The reused explanation omits this competing source of selling pressure.
\\
\midrule
\textbf{C. Depositor reassurance}\newline
21 March 2023. Janet Yellen pledges to protect depositors at smaller US banks amid withdrawals and banking-sector stress.
& \href{https://www.morganstanley.com/campaigns/wealth-management/move-report/move-report2023mar21/}{Morgan Stanley (2023)}\newline
Wealth Management Global Investment Office\newline
21 March 2023
& Links Yellen's remarks and other support measures to recovering confidence. The reused explanation does not address how the reassurance intervention could alter crisis-related safe-haven demand.
\\
\bottomrule
\end{tabular}
\caption{Contextual comparisons for three events sharing one normalized explanation. Normalization removes differences in capitalization and punctuation; raw explanations differ in terminal punctuation.}
\label{tab:reuse-context-cases}
\vspace{2pt}
\parbox{\textwidth}{\small\textit{Evidence scope.} These comparisons assess contextual justification. Case B addresses the broader sell-off; the World Gold Council also retains gold's longer-term safe-haven role. Case C provides a supplementary intervention-context comparison.}
\end{table}

\subsection{Polarity--Explanation Inconsistency}
\label{app:polarity-inconsistency}

Table~\ref{tab:polarity-explanation-cases} presents six cases in which the explanation's stated direction conflicts with the assigned polarity. These comparisons assess the internal directional consistency of generated graphs and their explanations.

\begin{table}[t]
\centering
\fontsize{10pt}{12pt}\selectfont
\setlength{\tabcolsep}{6pt}
\renewcommand{\arraystretch}{1.15}
\begin{tabular}{@{}>{\raggedright\arraybackslash}p{\dimexpr0.27\textwidth-1.5\tabcolsep\relax}
                  >{\raggedright\arraybackslash}p{\dimexpr0.21\textwidth-1.5\tabcolsep\relax}
                  >{\raggedright\arraybackslash}p{\dimexpr0.38\textwidth-1.5\tabcolsep\relax}
                  >{\raggedright\arraybackslash}p{\dimexpr0.14\textwidth-1.5\tabcolsep\relax}@{}}
\toprule
\textbf{Event and date} & \textbf{Model and edge} & \textbf{Generated explanation} & \textbf{Polarity} \\
\midrule
Jeremy Hunt refuses to rule out sanctions against China.\newline
4 July 2019
& \mbox{Llama-3.1-8B}\newline
\mbox{EVENT$\rightarrow$UUP}
& ``\textbf{US dollar strengthens} as investors seek safe-haven assets.''
& negative\newline
($\downarrow$)
\\
\midrule
US Federal Reserve raises short-term interest rates again.\newline
26 September 2018
& \mbox{Llama-3.1-8B}\newline
\mbox{EVENT$\rightarrow$VIX}
& ``Interest rate hike \textbf{increases expected volatility}''
& negative\newline
($\downarrow$)
\\
\midrule
US, Mexico and Canada reach a new trade deal.\newline
1 October 2018
& \mbox{Llama-3.1-8B}\newline
\mbox{EVENT$\rightarrow$VIX}
& ``\textbf{Market volatility increases} due to trade deal uncertainty''
& negative\newline
($\downarrow$)
\\
\midrule
Russia is accused of a cyber-attack on the chemical weapons watchdog.\newline
4 October 2018
& \mbox{Llama-3.1-8B}\newline
\mbox{EVENT$\rightarrow$VIX}
& ``Cyber-attack on chemical weapons watchdog \textbf{increases uncertainty and volatility}''
& negative\newline
($\downarrow$)
\\
\midrule
Fed chairman hints at the first interest rate cut in over a decade.\newline
10 July 2019
& \mbox{Qwen3-8B}\newline
\mbox{EVENT$\rightarrow$VIX}
& ``Market uncertainty about rate cut timing \textbf{increases volatility}''
& negative\newline
($\downarrow$)
\\
\midrule
Brexit trade talks may continue beyond the MEPs' deadline.\newline
19 December 2020
& \mbox{Llama-3.3-70B}\newline
\mbox{EVENT$\rightarrow$UUP}
& ``The event \textbf{may lead to a stronger U.S. dollar} due to increased uncertainty and potential for safe-haven demand.''
& negative\newline
($\downarrow$)
\\
\bottomrule
\end{tabular}
\caption{Selected cases of polarity--explanation inconsistency. All edges are direct EVENT-to-indicator relations: negative polarity specifies a downward target effect, whereas the explanations describe an upward effect. Event descriptions are shortened; explanations are reproduced verbatim, with directional phrases in bold.}
\label{tab:polarity-explanation-cases}
\end{table}

\section{Additional Result Visualization}
\label{app: add-result}

\paragraph{Full paired evaluation results}
Table~\ref{tab:full_paired_results} reports the matched time-series-only and graph-conditioned scores, together with their relative changes, across all evaluation metrics for each graph-generating LLM.

\begin{table}[t]
\centering
\fontsize{10pt}{12pt}\selectfont
\setlength{\tabcolsep}{4pt}
\renewcommand{\arraystretch}{1.15}

\begin{tabular}{@{}lllccccc@{}}
\toprule
& & &
\multicolumn{2}{c}{
\shortstack{\textbf{Reconstruction}\\\textbf{Fidelity}}}
&
\multicolumn{3}{c}{
\shortstack{\textbf{Market Distribution}\\\textbf{Consistency}}}
\\
\cmidrule(lr){4-5}
\cmidrule(lr){6-8}
\textbf{Category}
& \textbf{Model}
& \textbf{Condition}
& $\mu_d$
& $\sigma_d$
& $D_\mu$
& $D_b$
& $D_{\mathrm{dist}}$
\\
\midrule

\multirow{6}{*}{Proprietary}
& \multirow{3}{*}{GPT-5}
& TS-only
& 0.166577 & 0.155001 & 0.002574 & 0.007270 & 0.008075
\\
& & TS+Graph
& 0.173159 & 0.158475 & 0.002515 & 0.007045 & 0.007881
\\
& & $\Delta$ (\%)
& $+3.95$ & $+2.24$ & $-2.28$ & $-3.10$ & $-2.40$
\\
\cmidrule(lr){2-8}

& \multirow{3}{*}{Gemini-3.6-Flash}
& TS-only
& 0.178335 & 0.165734 & 0.002541 & 0.007252 & 0.008034
\\
& & TS+Graph
& 0.179873 & 0.162173 & 0.002625 & 0.007724 & 0.008526
\\
& & $\Delta$ (\%)
& $+0.86$ & $-2.15$ & $+3.31$ & $+6.51$ & $+6.12$
\\
\midrule

\multirow{12}{*}{Open-Weight}
& \multirow{3}{*}{Qwen3-8B}
& TS-only
& 0.153873 & 0.148719 & 0.002631 & 0.007302 & 0.008135
\\
& & TS+Graph
& 0.159867 & 0.153475 & 0.002680 & 0.007585 & 0.008406
\\
& & $\Delta$ (\%)
& $+3.90$ & $+3.20$ & $+1.87$ & $+3.88$ & $+3.34$
\\
\cmidrule(lr){2-8}

& \multirow{3}{*}{Qwen3-32B}
& TS-only
& 0.158514 & 0.148282 & 0.002630 & 0.007272 & 0.008101
\\
& & TS+Graph
& 0.163781 & 0.153941 & 0.002683 & 0.007274 & 0.008110
\\
& & $\Delta$ (\%)
& $+3.32$ & $+3.82$ & $+1.99$ & $+0.03$ & $+0.10$
\\
\cmidrule(lr){2-8}

& \multirow{3}{*}{Llama-3.1-8B}
& TS-only
& 0.161483 & 0.144823 & 0.002650 & 0.007282 & 0.008116
\\
& & TS+Graph
& 0.168763 & 0.154176 & 0.002768 & 0.007517 & 0.008413
\\
& & $\Delta$ (\%)
& $+4.51$ & $+6.46$ & $+4.45$ & $+3.23$ & $+3.65$
\\
\cmidrule(lr){2-8}

& \multirow{3}{*}{Llama-3.3-70B}
& TS-only
& 0.206081 & 0.192130 & 0.002298 & 0.007273 & 0.007950
\\
& & TS+Graph
& 0.198407 & 0.188260 & 0.002243 & 0.007865 & 0.008499
\\
& & $\Delta$ (\%)
& $-3.72$ & $-2.01$ & $-2.39$ & $+8.14$ & $+6.89$
\\

\bottomrule
\end{tabular}

\captionsetup{font={10pt,rm}}
\caption{
Full paired results for the main evaluation.
TS-only receives historical observations, while TS+Graph
receives those observations plus the original generated graph. Relative change is computed as
$\Delta = 100 \times
(\text{Graph-conditioned} - \text{Historical-only})/
\text{Historical-only}$.
Lower scores are better; negative changes indicate improvements.
Confidence intervals for relative changes are reported in Table~\ref{tab:main_results};
}
\label{tab:full_paired_results}
\end{table}

\paragraph{VAR calibration results}
Table~\ref{tab:var-calibration} summarizes the relative changes for VAR calibration.
The full results below report absolute scores under TS-only, reference-graph,
node-shuffled reference, and event-mismatched
reference conditions across all five evaluation
metrics. For each graph-conditioned condition,
we also report relative changes from the matched
TS-only baseline and paired-bootstrap 95\%
confidence intervals.

\begin{table}[t]
\centering
\fontsize{9pt}{10.8pt}\selectfont
\setlength{\tabcolsep}{5pt}
\renewcommand{\arraystretch}{1.25}

\newcommand{\varfullcell}[3]{%
\shortstack{$#1$\\{\scriptsize $[#2,#3]$}}%
}

\begin{adjustbox}{max width=\textwidth}
\begin{tabular}{lccccc}
\toprule
& \multicolumn{2}{c}{\textbf{Reconstruction Fidelity}}
& \multicolumn{3}{c}{\textbf{Market Distribution Consistency}}
\\
\cmidrule(lr){2-3}
\cmidrule(lr){4-6}

\textbf{Graph input}
& $\mu_d$
& $\sigma_d$
& $D_\mu$
& $D_b$
& $D_{\mathrm{dist}}$
\\
\midrule

\multicolumn{6}{l}{\textit{Absolute scores}}
\\

TS-only
& 1.175745 & 0.926263 & 0.002783 & 0.008963 & 0.009697
\\

VAR reference
& 1.016284 & 0.839030 & 0.002452 & 0.008622 & 0.009240
\\

Node-shuffled
& 1.317926 & 1.103682 & 0.002757 & 0.008981 & 0.009770
\\

Event-mismatched
& 1.165599 & 0.926356 & 0.002935 & 0.009066 & 0.009839
\\
\midrule

\multicolumn{6}{l}{
\textit{Relative changes from TS-only (\%)
with 95\% confidence intervals}}
\\

VAR reference
& \varfullcell{-13.56}{-20.61}{-5.85}
& \varfullcell{-9.42}{-16.42}{-1.51}
& \varfullcell{-11.92}{-20.99}{-0.95}
& \varfullcell{-3.80}{-6.12}{-1.44}
& \varfullcell{-4.72}{-6.59}{-2.74}
\\

Node-shuffled
& \varfullcell{+12.09}{+2.63}{+22.92}
& \varfullcell{+19.15}{+8.62}{+30.86}
& \varfullcell{-0.93}{-13.14}{+14.23}
& \varfullcell{+0.20}{-2.62}{+3.05}
& \varfullcell{+0.74}{-1.72}{+3.42}
\\

Event-mismatched
& \varfullcell{-0.86}{-6.85}{+5.54}
& \varfullcell{+0.01}{-6.25}{+6.47}
& \varfullcell{+5.45}{-8.78}{+19.16}
& \varfullcell{+1.14}{-2.02}{+4.36}
& \varfullcell{+1.46}{-1.55}{+4.31}
\\

\bottomrule
\end{tabular}
\end{adjustbox}

\caption{
Full  VAR-reference results.
The upper panel reports mean event-level scores.
The lower panel reports relative percentage changes
from the matched TS-only baseline, with paired
event-bootstrap 95\% confidence intervals.
All conditions use the same events and forecasting horizons.
Lower absolute scores are better;
negative changes indicate improvements.
}
\label{tab:full-var-calibration}
\end{table}

\section{Details of Metric Definitions}
\label{app:evl_metric}




\paragraph{Reconstruction-error normalization}
To account for differences in scale across financial
indicators, reconstruction errors are normalized using
indicator-specific standard deviations computed from
a fixed reference period shared across all events.

For each indicator $I_j$, let
$\{x_j^{(\tau)}\}_{\tau=1}^{N}$ denote its $N=1393$
daily observations in the fixed reference period from
19 June 2018 to 31 December 2023. We define
\begin{equation}
\begin{aligned}
\bar{x}_j
&= \frac{1}{N}\sum_{\tau=1}^{N}x_j^{(\tau)},\\
s_j
&= \left[
\frac{1}{N}\sum_{\tau=1}^{N}
\left(x_j^{(\tau)}-\bar{x}_j\right)^2
\right]^{1/2}.
\end{aligned}
\end{equation}
We compute
$d_j^{(t)}=|\hat{y}_j^{(t)}-y_j^{(t)}|/s_j$
and summarize these standardized absolute errors
by $\mu_d$ and $\sigma_d$.

The same indicator-specific scales are used across all events, graph-generating models, and evaluation conditions. All 20 reference standard deviations are strictly positive,
with a minimum of approximately 1.49; no denominator clipping is applied.

\paragraph{Log Returns}
Before computing log returns, both observed and predicted
values are lower-bounded by $\epsilon=10^{-8}$ to ensure
that logarithms are well-defined. For each financial indicator $I_j$, the observed log return at time step $t$ is defined as

\begin{equation}
r_j^{(t)}
=
\log\!\left(\max\{y_j^{(t)},\epsilon\}\right)
-
\log\!\left(\max\{y_j^{(t-1)},\epsilon\}\right),
\qquad t=2,\ldots,T.
\end{equation}
and the reconstructed log return $\hat{r}_j^{(t)}$ is defined analogously from $\hat{y}_j^{(t)}$. Returns are computed only between consecutive observations
within the future window, excluding the transition from
the last historical observation to the first future value.
Thus, distribution consistency assesses within-window
fluctuations rather than the initial change from history
to the future window. The first future value remains
included in the reconstruction-fidelity metrics for
graph-involved indicators. To characterize the overall market behavior, we aggregate the observed and reconstructed log returns across all 20 financial indicators and time steps:
\begin{equation}
\mathcal{R}
=
\left\{
r_j^{(t)}
\mid
j=1,\ldots,20,\;
t=2,\ldots,T
\right\},
\end{equation}
\begin{equation}
\hat{\mathcal{R}}
=
\left\{
\hat{r}_j^{(t)}
\mid
j=1,\ldots,20,\;
t=2,\ldots,T
\right\}.
\end{equation}

\paragraph{Laplace Distribution Estimation}
Following the heavy-tailed behavior commonly observed in financial returns, we model each aggregated return set using a Laplace distribution:
\begin{equation}
p(r\mid\mu,b)
=
\frac{1}{2b}
\exp\left(
-\frac{|r-\mu|}{b}
\right),
\end{equation}
where $\mu$ and $b$ denote the location and scale parameters, respectively. For the observed return set $\mathcal{R}$, these parameters are estimated using maximum-likelihood estimation:
\begin{equation}
\mu
=
\operatorname{median}(\mathcal{R}),
\qquad
b
=
\frac{1}{|\mathcal{R}|}
\sum_{r\in\mathcal{R}}
|r-\mu|.
\end{equation}
The resulting observed market-return distribution is
\begin{equation}
P_{\mathrm{real}}
=
\operatorname{Laplace}(\mu,b).
\end{equation}
The reconstructed parameters $(\hat{\mu},\hat{b})$ are estimated analogously from $\hat{\mathcal{R}}$, yielding
\begin{equation}
P_{\mathrm{rec}}
=
\operatorname{Laplace}(\hat{\mu},\hat{b}).
\end{equation}
The differences between these distributions are evaluated using $D_\mu$, $D_b$, and $D_{\mathrm{dist}}$, as defined in Section~\ref{sec:evaluation-protocol}.

\section{Graph Generation Prompt Template and Inference Details}
\label{app:graph-generation-prompts}

\subsection{Graph Generation Prompt Template}
\label{app:graph-prompt-template}
All models are evaluated using the same prompt template. For each event, we
instantiate the template with the event headline and trail text, the preceding
63 daily observations, and metadata for the available financial indicators.
The complete prompt template used for graph generation is shown in
Figure~\ref{fig:graph-generation-prompt}.

\subsection{Inference Settings}
All models generate one graph per event using the unified prompt template in
Appendix~\ref{app:graph-prompt-template}. We set the maximum propagation depth to
three steps and the maximum output length to 4,096 tokens for all models.
Open-weight models are served locally with vLLM using greedy decoding
(temperature $=0$). GPT-5 and Gemini-3.6-Flash use their provider-supported
reasoning configurations. Provider-supported structured JSON output is used
where available, and all outputs are subsequently processed using the same
parser and graph validator.

\begin{table}[t]
\centering
\fontsize{9pt}{10.8pt}\selectfont
\setlength{\tabcolsep}{3pt}
\begin{tabular}{@{}>{\raggedright\arraybackslash}p{\dimexpr0.40\columnwidth-4pt\relax}>{\raggedright\arraybackslash}p{\dimexpr0.27\columnwidth-4pt\relax}>{\raggedright\arraybackslash}p{\dimexpr0.33\columnwidth-4pt\relax}@{}}
\toprule
\textbf{Model} & \textbf{Access} & \textbf{Inference setting} \\
\midrule
GPT-5
& OpenAI Responses API
& Reasoning effort: low \\

Gemini-3.6-Flash
& Gemini API
& Thinking level: low \\

Qwen3-8B-Instruct
& Local vLLM
& Temperature $=0$ \\

Qwen3-32B-Instruct
& Local vLLM
& Temperature $=0$ \\

Llama-3.1-8B-Instruct
& Local vLLM
& Temperature $=0$ \\

Llama-3.3-70B-Instruct
& Local vLLM
& Temperature $=0$ \\
\bottomrule
\end{tabular}
\caption{Model-specific graph-generation inference settings. All models use
the same prompt template, a maximum of three propagation steps, and a maximum
output length of 4,096 tokens.}
\label{tab:graph-inference-settings}
\end{table}

\begingroup
\captionsetup{hypcap=false}
\begin{tcolorbox}[breakable,width=\columnwidth,colback=white,colframe=black,boxrule=0.4pt,arc=0pt,left=3pt,right=3pt,top=3pt,bottom=3pt]
\fontsize{9pt}{10.8pt}\selectfont
\begin{Verbatim}[breaklines=true,breakanywhere=true,breaksymbolleft={},fontsize=\fontsize{9pt}{10.8pt}\selectfont]
# Task

Given an event description and preceding historical time-series data, generate
a directed event-induced causal graph over the financial indicators affected
by the event.

# Input

Here is the event to be represented by `EVENT` node in the graph:

<event>
Headline: {{headline}}
Summary: {{trail_text}}
</event>

Here are the historical daily time series preceding the event. They
describe the market context before the event:

<historical_time_series>
{{time_series_csv}}
</historical_time_series>

Here are the available financial indicators that may be used as graph nodes:

<available_indicators>
{{indicator_metadata}}
</available_indicators>

# Output Format

Return the result in JSON format only. Do not include any extra text.

```json
{
  "duration_days": <7 or 15>,
  "step1": [
    {
      "source": "EVENT",
      "target": "INDICATOR_NAME",
      "polarity": "<positive or negative>",
      "explanation": "economic mechanism explanation"
    },
    ...
  ],
  "step2": [
    {
      "source": "INDICATOR_NAME",
      "target": "INDICATOR_NAME",
      "polarity": "<positive or negative>",
      "explanation": "economic mechanism explanation"
    },
    ...
  ],
  "step3": [
    {
      "source": "INDICATOR_NAME",
      "target": "INDICATOR_NAME",
      "polarity": "<positive or negative>",
      "explanation": "economic mechanism explanation"
    },
    ...
  ]
}
```

`EVENT` is the event node. When an edge is included, replace
`INDICATOR_NAME` with one name from the available_indicators.
`...` denotes zero or more additional edge objects and must not be output
literally.

# Requirements

1. Select `duration_days` as either 7 or 15, representing how long the
   event-induced impact is expected to remain relevant.
2. `step1` contains edges from `EVENT` to indicators directly affected by the
   event. If further effects are identified, `step2` and `step3` contain edges
   from indicators affected in the previous step: every source in `step2` must
   be a target from `step1`, and every source in `step3` must be a target from
   `step2`.
3. Include only edges through which the event's effects continue to propagate.
   Stop when no further propagation is identified; omit all later steps.
4. Do not include duplicate edges or cycles. Give each edge a concise
   explanation of its economic mechanism.
5. Assign each included edge a `polarity` of either `positive` or `negative`.
   For `step1`, `positive` indicates an upward direct effect of the event on
   the `target`, and `negative` indicates a downward direct effect. For
   `step2` and `step3`, `positive` indicates that the effect transmitted
   through the edge to the `target` is in the same direction as the
   event-induced change in the `source`, and `negative` indicates the
   opposite direction.
\end{Verbatim}
\end{tcolorbox}
\captionof{figure}{Unified graph-generation prompt template used for all evaluated
models. The placeholders are instantiated separately for each event.}
\label{fig:graph-generation-prompt}
\endgroup














\section{Forecasting Evaluator Implementation}
\label{app: evaluator-implementation}

\subsection{Evaluator and Inference Settings}

We use \texttt{Qwen3.8-27B} ~\cite{qwen38} as a fixed zero-shot forecasting evaluator for all graph-generating LLMs. The evaluator receives 63 historical daily observations for 20 financial indicators, represented by anonymous indices from \texttt{index\_0} to \texttt{index\_19}. The same index mapping is used in the historical time series, the generated graph, and the forecast output. Raw time-series values are rounded to four decimal places when serialized into the prompt.

For each event, we evaluate two matched conditions: (1) a time-series-only condition and (2) a graph-conditioned condition that additionally receives the generated event-induced graph. Both conditions use the duration $d\in\{7,15\}$ selected by the graph-generating LLM. No examples or future target values are provided to the evaluator. The complete inference settings are summarized in Table~\ref{tab:evaluator-inference-settings}.

\begin{table}[t]
\centering
\fontsize{9pt}{10.8pt}\selectfont
\setlength{\tabcolsep}{3pt}
\begin{tabular}{@{}>{\raggedright\arraybackslash}p{\dimexpr0.44\linewidth-3pt\relax}>{\raggedright\arraybackslash}p{\dimexpr0.56\linewidth-3pt\relax}@{}}
\toprule
\textbf{Setting} & \textbf{Value} \\
\midrule
Evaluator & Qwen3.8-27B \\
Evaluation mode & Zero-shot \\
Thinking mode & Disabled \\
Decoding & Greedy decoding \\
Temperature & 0 \\
Maximum context length & 24,576 tokens \\
Maximum output length & 12,000 tokens \\
Historical window & 63 daily observations \\
Forecasting horizon & 7 or 15 daily steps \\
Time-series representation & Raw values, rounded to four decimals \\
Inference framework & vLLM \\
Tensor parallelism & 2 GPUs \\
\bottomrule
\end{tabular}
\caption{Inference settings for the fixed forecasting evaluator.}
\label{tab:evaluator-inference-settings}
\end{table}

\subsection{Time-series Only Prompt Template and Graph-Conditioned Template}
\label{app: evaluator-prompt}

Figures~\ref{fig:evaluator-ts-only-prompt}
and~\ref{fig:evaluator-graph-prompt} present the time-series-only and
graph-conditioned evaluator prompt templates, respectively. The two evaluator prompts share the same task description, historical
time series, forecasting horizon, and output requirements. The
graph-conditioned prompt differs only by additionally providing the generated
event-induced graph. The placeholders \texttt{\{d\}}, \texttt{\{history\}},
and \texttt{\{graph\}} are instantiated separately for each event.

\begingroup
\captionsetup{hypcap=false}
\begin{tcolorbox}[breakable,width=\columnwidth,colback=white,colframe=black,boxrule=0.4pt,arc=0pt,left=3pt,right=3pt,top=3pt,bottom=3pt]
\fontsize{9pt}{10.8pt}\selectfont
\begin{Verbatim}[breaklines=true,breakanywhere=true,breaksymbolleft={},fontsize=\fontsize{9pt}{10.8pt}\selectfont]
I have a multivariate time-series forecasting task for you.

Given the historical time series of 20 indexed variables and the auxiliary
causal graph encoding an external event's effects below, predict the next
{d} daily values for all indexed variables.

Here is the historical time series. `days` lists the 63 historical daily
steps. For each indexed variable, `values_by_day[i]` is its value at
`days[i]`.

<history>
{history}
</history>

Here is an auxiliary event-induced causal graph representing the
event-specific causal structure over the 20 indexed variables during the
forecasting window, in addition to the dependencies reflected in the
historical time series. The graph encodes only the structure and polarity
of the event-induced effects:
- `event_to_indices`: Each entry specifies a `target` index directly
  affected by the event. Its `polarity` describes the direction of this
  direct effect: `positive` indicates an upward effect on the `target`,
  and `negative` indicates a downward effect.
- `propagation_edges`: Each entry represents a directed relation through
  which the event-induced effects subsequently propagate from its
  `source` index to its `target` index. Its `polarity` indicates whether
  the effect transmitted to the `target` is in the same direction
  (`positive`) or the opposite direction (`negative`) as the event-induced
  change in the `source`.

<graph>
{graph}
</graph>

Return only the forecast JSON between <forecast> and </forecast> tags.
Include exactly one `values_by_day` list for every index from `index_0`
through `index_19`; each list must contain its next {d} daily values.
Do not include `days` or any other information.

<forecast>
{
  "time_series": {
    "index_0": {"values_by_day": [value_1, ..., value_{d}]},
    ...
    "index_19": {"values_by_day": [value_1, ..., value_{d}]}
  }
}
</forecast>
\end{Verbatim}
\end{tcolorbox}
\captionof{figure}{Graph-conditioned forecasting prompt template used by the evaluator.}
\label{fig:evaluator-graph-prompt}
\endgroup

\begingroup
\captionsetup{hypcap=false}
\begin{tcolorbox}[breakable,width=\columnwidth,colback=white,colframe=black,boxrule=0.4pt,arc=0pt,left=3pt,right=3pt,top=3pt,bottom=3pt]
\fontsize{9pt}{10.8pt}\selectfont
\begin{Verbatim}[breaklines=true,breakanywhere=true,breaksymbolleft={},fontsize=\fontsize{9pt}{10.8pt}\selectfont]
I have a multivariate time-series forecasting task for you.

Given the historical time series of 20 indexed variables, predict the next
{d} daily values for all indexed variables.

Here is the historical time series. `days` lists the 63 historical daily
steps. For each indexed variable, `values_by_day[i]` is its value at
`days[i]`.

<history>
{history}
</history>

Return only the forecast JSON between <forecast> and </forecast> tags.
Include exactly one `values_by_day` list for every index from `index_0`
through `index_19`; each list must contain its next {d} daily values.
Do not include `days` or any other information.

<forecast>
{
  "time_series": {
    "index_0": {"values_by_day": [value_1, ..., value_{d}]},
    ...
    "index_19": {"values_by_day": [value_1, ..., value_{d}]}
  }
}
</forecast>
\end{Verbatim}
\end{tcolorbox}
\captionof{figure}{Time-series-only forecasting prompt template used by the evaluator.}
\label{fig:evaluator-ts-only-prompt}
\endgroup

\subsection{Input and Output Format}

The historical observations are serialized as indexed JSON:

\begin{verbatim}
{
  "days": ["day_1", ..., "day_63"],
  "time_series": {
    "index_0": {
      "values_by_day": [x_0^1, ..., x_0^63]
    },
    ...
    "index_19": {
      "values_by_day": [x_19^1, ..., x_19^63]
    }
  }
}
\end{verbatim}

The generated graph is converted into the following schema:

\begin{verbatim}
{
  "event_to_indices": [
    {"target": "index_0", "polarity": "positive"},
    {"target": "index_3", "polarity": "negative"}
  ],
  "propagation_edges": [
    {
      "source": "index_0",
      "target": "index_5",
      "polarity": "positive"
    },
    {
      "source": "index_5",
      "target": "index_8",
      "polarity": "negative"
    }
  ]
}
\end{verbatim}

Each entry in \texttt{event\_to\_indices} specifies
a directly affected indicator and the polarity
of the event's effect. Each entry in
\texttt{propagation\_edges} specifies a source,
a target, and whether the transmitted effect
follows or reverses the direction of the
event-induced change in the source. Indicator
names are omitted from both evaluator conditions;
the index mapping remains fixed across the
historical series, graph, and forecast.

The evaluator is required to return one forecast sequence for every indicator:

\begin{verbatim}
<forecast>
{
  "time_series": {
    "index_0": {
      "values_by_day": [y_0^1, ..., y_0^d]
    },
    ...
    "index_19": {
      "values_by_day": [y_19^1, ..., y_19^d]
    }
  }
}
</forecast>
\end{verbatim}

\subsection{Duration Routing and Score Aggregation}

For each event, the duration predicted by the graph-generating LLM determines whether the evaluator produces a 7-day or 15-day forecast. The same duration is used for the matched time-series-only and graph-conditioned conditions. Thus, duration selection is held fixed within each paired comparison.

Each metric is first computed separately for each event over its selected forecasting window. Event-level scores are then averaged across all valid paired events, so that 7-day and 15-day events receive equal weight in the overall result. 

\subsection{Graph Control Construction}
\label{app:graph-controls}

We construct two controls for each LLM-generated
graph to examine how reconstruction depends on
the encoded relationships and their relevance to
the event being evaluated. Both controls are
provided to the evaluator in the same anonymized
signed-graph format as the original graph.

\paragraph{Node-shuffled control}
We permute indicator identities within the set
of indicators involved in the original graph,
while keeping the event node and graph topology
unchanged. For graphs
with at least two indicator nodes, the permutation
ensures that no indicator retains its original
assignment.  We then independently assign positive or negative
polarity to each edge with equal probability,
including direct event-to-indicator edges.
This control therefore perturbs both indicator
assignments and directional information.

\paragraph{Event-mismatched control}
We replace each original graph with one generated
by the same LLM for another event with the same
forecasting duration. Graphs are reassigned
one-to-one. We exclude graphs whose connections
are identical to those of the original graph
when edge polarity is ignored. The replacement
graph retains its original edges and polarities.

Only the graph input changes across conditions;
historical observations, forecasting targets,
and durations remain fixed for each event.
All comparisons use the same valid events.
Reconstruction fidelity is computed over the
indicators involved in the original graph,
while market-distribution metrics use all
20 indicators.

\subsection{Paired Bootstrap Confidence Intervals}
\label{app:paired-ci}

We report 95\% confidence intervals using a paired
bootstrap over events. For each comparison, we
retain events with valid forecasts in all compared
conditions and average event-level scores with
equal weights. The relative change from TS only is
\begin{equation}
\Delta M_c =
100 \times
\frac{\overline{m}_c-\overline{m}_{\mathrm{TS}}}
     {\overline{m}_{\mathrm{TS}}},
\end{equation}
where $\overline{m}_c$ and
$\overline{m}_{\mathrm{TS}}$ are the mean scores
for condition $c$ and the TS-only baseline.

We draw 5,000 bootstrap samples, each containing
the original number of events sampled with
replacement. The same sampled events are used
for both conditions, and $\Delta M_c$ is
recomputed for each sample. The 2.5th and 97.5th
percentiles form the 95\% confidence interval.


\section{VAR Reference Graph Construction and Evaluation}
\label{app:var-calibration}

We construct future-informed statistical reference
graphs to calibrate the forecasting evaluator.
The construction uses 600 fitting events and
150 evaluation events with disjoint publication
dates. Separate dependency matrices and reference
graphs are constructed for 7- and 15-day horizons.

\paragraph{Reference signals and dependency estimation}
For each event, we fit a ridge-regularized VAR(1)
model to log returns computed from its 63
historical daily observations, with ridge penalty
1.0. We derive two post-event signals:
$u_{e,t}$, the observed return minus its one-step
VAR prediction using the preceding observed return;
and $\delta_{e,t}$, the observed return minus the
VAR prediction obtained by recursively forecasting
from the last historical return. Both signals are
standardized by the corresponding indicator's
historical VAR residual standard deviation,
yielding $z_{u,e,t}$ and $z_{\delta,e,t}$.

For each horizon, we fit no-intercept ElasticNet
regressions from $z_{\delta,e,t-1}$ to
$z_{u,e,t}$ using only the fitting events.
The L1 mixing ratio is 0.8. Regularization is
selected by five-fold cross-validation grouped
by publication date, prioritizing candidates
with an off-diagonal coefficient density between
10\% and 25\%. The selected regularization
strength is 0.4 for both horizons.
We retain the off-diagonal coefficients as
the dependency matrix $C$.

\paragraph{Graph construction and polarity}
For each evaluation event, the two indicators
with the largest absolute first-day innovations
$|z_{u,e,j,1}|$ become direct targets of
\texttt{EVENT}. For a candidate propagation
edge $i\rightarrow j$, we use the score
\begin{equation}
s_{e,i\rightarrow j}
=
\max_{a_i<t\leq d}
\left|
C_{ji}\,
z_{\delta,e,i,t-1}\,
z_{u,e,j,t}
\right|,
\end{equation}
where $a_i$ is the selected arrival day of the
source's effect and $d$ is the forecasting
horizon. Direct targets have $a_i=1$; each
subsequent target inherits the day attaining
its selected edge's maximum score.

We select up to two propagation edges in the
second layer and one in the third, ranked by
this score among nonzero coefficients. Sources
must belong to the preceding layer, and targets
must be previously unselected indicators.
These constraints preserve temporal ordering
and give each indicator at most one incoming
edge. Direct-edge polarity follows the sign
of $z_{u,e,j,1}$, while propagation-edge polarity
follows the sign of $C_{ji}$.

\paragraph{Control and evaluation}
We construct node-shuffled reference graphs
using the procedure in
Appendix~\ref{app:graph-controls}, preserving
the involved indicator set and topology while
permuting indicator identities and independently
randomizing edge polarities. Each event is randomly assigned a forecasting
duration of 7 or 15 days before reference-graph
construction. The 150 evaluation events are
assessed under TS-only, reference-graph, and
node-shuffled conditions using their assigned
durations, which are held fixed across conditions.
Results are averaged with equal event weights. We use the fixed Qwen3.8-27B evaluator
and the prompt and inference settings in
Appendix~\ref{app: evaluator-implementation}.
Reconstruction fidelity is computed over the
original reference graph's indicators, and
market-distribution metrics use all 20 indicators.
Confidence intervals follow
Appendix~\ref{app:paired-ci}.

\section{Graph Specificity and Explanation Similarity}
\label{app:graph-specificity}

This appendix provides the detailed definitions and complete results for the graph specificity and explanation similarity analyses described in Section~\ref{sec:graph-specificity}. All analyses are conducted separately for direct event-to-indicator edges and indicator-to-indicator propagation edges.

\subsection{Relation-level Grouping}

For each event $e$ and edge type
$k\in\{\mathrm{dir},\mathrm{prop}\}$, let $E_e^{(k)}$ denote the generated edge set. Each edge is represented by a directed relation $r=(u,v)$ and has an associated explanation $x_{e,r}$. For each relation $r$, we group all of its occurrences across events:

\begin{equation}
\mathcal{X}^{(k)}_r
=
\left\{
(e,x_{e,r})
\mid
r\in E_e^{(k)}
\right\}.
\end{equation}

Thus, the elements being grouped are event--edge instances, while the grouping key is the same directed relation $r$. For direct edges, $r=(\mathrm{EVENT},v)$; for propagation edges, $r=(u,v)$ for two financial indicators. Let

\begin{equation}
n_r = |\mathcal{X}^{(k)}_r|
\end{equation}

denote the number of occurrences of relation $r$, and let $\mathcal{R}^{(k)}$ denote the set of distinct relations of edge type $k$.

\subsection{Structural Concentration}

For each edge type $k$, structural concentration is measured by the proportion of all generated edges represented by the three most frequent relations:

\begin{equation}
C^{(k)}_{\mathrm{top3}}
=
\frac{
\sum_{r\in\mathrm{Top3}^{(k)}} n_r
}{
\sum_{r\in\mathcal{R}^{(k)}} n_r
},
\end{equation}

where $\mathcal{R}^{(k)}$ is the set of distinct relations of type $k$, and $\mathrm{Top3}^{(k)}$ contains the three relations with the largest occurrence counts. Higher values indicate that a small set of relations accounts for a larger proportion of the generated edges.

\subsection{Explanation Similarity}

For each relation $r$, we compute standard Self-BLEU-4 \citep{zhu2018texygen} over the explanations in $\mathcal{X}^{(k)}_r$. Each explanation is treated as a candidate, with the remaining explanations for the same relation across events used as references. We include relations occurring in at least five events and report the macro average:

\begin{equation}
S^{(k)}_{\mathrm{SB}}
=
\frac{1}{
|\mathcal{R}^{(k)}_{\geq 5}|
}
\sum_{r\in\mathcal{R}^{(k)}_{\geq 5}}
\mathrm{SelfBLEU}_4
\left(
\mathcal{X}^{(k)}_r
\right).
\end{equation}

Before scoring, explanations are lowercased and tokenized into alphanumeric tokens. At most 60 explanations are deterministically sampled from each relation group. Higher values indicate greater textual similarity among explanations of the same relation across events.

\subsection{Exact Explanation Reuse}

To determine whether explanation similarity includes identical text reuse, we additionally compute exact explanation reuse. For each relation $r$, let $u_r$ denote the number of distinct normalized explanation strings in $\mathcal{X}^{(k)}_r$. Explanations are normalized by lowercasing and removing punctuation and redundant whitespace. The pooled exact explanation reuse rate is:

\begin{equation}
R^{(k)}_{\mathrm{pooled}}
=
1-
\frac{
\sum_{r\in\mathcal{R}^{(k)}}u_r
}{
\sum_{r\in\mathcal{R}^{(k)}}n_r
}.
\end{equation}

A higher value indicates that a larger proportion of explanation instances repeat exactly the text previously used for the same relation.

As a robustness check, we also report the macro exact reuse rate, which assigns equal weight to each relation appearing in at least five events:

\begin{equation}
R^{(k)}_{\mathrm{macro}}
=
\frac{1}{
|\mathcal{R}^{(k)}_{\geq 5}|
}
\sum_{r\in\mathcal{R}^{(k)}_{\geq 5}}
\left(
1-\frac{u_r}{n_r}
\right).
\end{equation}

\subsection{Full Results}

Table~\ref{tab:full-graph-specificity} reports structural concentration and relation-conditioned explanation similarity for all evaluated LLMs. Table~\ref{tab:full-exact-reuse} reports pooled and macro exact explanation reuse. The pooled measure reflects repetition in the complete set of generated explanations, whereas the macro measure tests whether reuse remains after assigning equal weight to eligible relation groups. The Llama models exhibit the highest exact reuse under both measures, indicating that their repeated explanations are not solely attributable to a single high-frequency relation.


\begin{table}[t]
\centering
\fontsize{9pt}{10.8pt}\selectfont
\setlength{\tabcolsep}{2pt}
\begin{tabular*}{\columnwidth}{@{\extracolsep{\fill}}lrrrr@{}}
\toprule
& \multicolumn{2}{c}{Top-3 mass (\%)} & \multicolumn{2}{c}{Self-BLEU-4} \\
\cmidrule(lr){2-3}\cmidrule(lr){4-5}
Model & D & P & D & P \\
\midrule
\multicolumn{5}{c}{\textit{Open-weight Models}} \\
Llama-3.1-8B & \textbf{97.97} & 44.93 & 0.488 & \textbf{0.687} \\
Llama-3.3-70B & 76.24 & 24.60 & \textbf{0.512} & 0.667 \\
Qwen3-8B & 86.81 & 36.48 & 0.452 & 0.679 \\
Qwen3-32B & 45.98 & 18.85 & 0.489 & 0.574 \\
\midrule
\multicolumn{5}{c}{\textit{Proprietary Models}} \\
GPT-5 & 40.29 & 23.95 & 0.416 & 0.472 \\
Gemini-3.6-Flash & 83.27 & \textbf{58.10} & 0.239 & 0.434 \\
\bottomrule
\end{tabular*}
\caption{Full structural concentration and relation-conditioned explanation similarity results.
Top-3 mass is reported as a percentage; higher Self-BLEU indicates greater
textual similarity among explanations for the same relation across events. D: Direct; P: Propagation.}
\label{tab:full-graph-specificity}
\end{table}


\begin{table}[t]
\centering
\fontsize{9pt}{10.8pt}\selectfont
\setlength{\tabcolsep}{2pt}
\begin{tabular*}{\columnwidth}{@{\extracolsep{\fill}}lrrrr@{}}
\toprule
& \multicolumn{2}{c}{\shortstack{\textbf{Pooled}\\\textbf{Exact Reuse}}}
& \multicolumn{2}{c}{\shortstack{\textbf{Macro Exact Reuse}\\($n_r\geq5$)}} \\
\cmidrule(lr){2-3}\cmidrule(lr){4-5}
\textbf{Model} & D & P & D & P \\
\midrule
\multicolumn{5}{c}{\textit{Open-weight Models}} \\
Llama-3.1-8B & \textbf{58.13} & \textbf{64.64} & \textbf{30.70} & \textbf{31.68} \\
Llama-3.3-70B & 42.71 & 34.42 & 17.56 & 13.37 \\
Qwen3-8B & 17.76 & 28.05 & 6.84 & 9.10 \\
Qwen3-32B & 7.07 & 7.32 & 2.80 & 1.77 \\
\midrule
\multicolumn{5}{c}{\textit{Proprietary Models}} \\
GPT-5 & 12.50 & 13.36 & 5.73 & 3.91 \\
Gemini-3.6-Flash & 0.05 & 5.04 & 0.01 & 1.05 \\
\bottomrule
\end{tabular*}
\caption{Exact explanation reuse (\%). Pooled reuse is edge-weighted across
all relation groups. Macro reuse assigns equal weight to relations appearing
in at least five events. D: Direct; P: Propagation.}
\label{tab:full-exact-reuse}
\end{table}

\end{document}